\documentclass[nonacm]{acmart}

\usepackage{natbib}
\usepackage{amsmath}
\usepackage{amsthm}
\usepackage{xspace}
\usepackage{algorithm2e}
\usepackage{graphicx}
\usepackage{adjustbox}
\usepackage{subfig}
\usepackage{hyperref, hypernat, url}
\usepackage{glossaries}
\usepackage{verbatim}
\usepackage{tikz}
\usepackage{pgfplots}
\usepackage{booktabs}
\usepackage{makecell}
\usepackage{paralist}
\usepackage{cleveref}
\usepackage{nicefrac}
\usetikzlibrary{arrows.meta,graphs,shapes.misc,calc}

\definecolor{beaublue}{rgb}{0.74, 0.83, 0.9}
\definecolor{coralpink}{rgb}{0.97, 0.51, 0.47}
\definecolor{grannysmithapple}{rgb}{0.66, 0.89, 0.63}
\definecolor{magicmint}{rgb}{0.67, 0.94, 0.82}

\usepackage[]{todonotes}

\LetLtxMacro{\todonote}{\todo}
\renewcommand{\todo}[2][]
{\todonote[inline, caption={#2}, size=\footnotesize, #1]
{\renewcommand{\baselinestretch}{0.5}\selectfont#2\par}}

\theoremstyle{definition}
\newtheorem{definition}{Definition}

\theoremstyle{plain}

\newtheorem{lemma}{Lemma}

\RestyleAlgo{ruled}

\renewcommand{\Pr}[1]{\mathrm{Pr}\left[ #1 \right]}

\newcommand{\eg}{e.g.\xspace}
\newcommand{\ie}{i.e.\xspace}

\crefformat{enumi}{(#1)}

\title{Rethinking Privacy in Machine Learning Pipelines from an Information Flow Control Perspective}

\author{Lukas Wutschitz}
\affiliation{
  \institution{M365 Research}
  \city{Cambridge}
  \country{UK}
}

\author{Boris K\"opf}
\email{boris.koepf@microsoft.com}
\affiliation{
  \institution{Azure Research}
  \city{Cambridge}
  \country{UK}
}

\author{Andrew Paverd}
\email{andrew.paverd@microsoft.com}
\affiliation{
  \institution{Microsoft Security Response Center}
  \city{Cambridge}
  \country{UK}
}

\author{Saravan Rajmohan}
\email{saravan.rajmohan@microsoft.com}
\affiliation{
  \institution{M365 Research}
  \city{Cambridge}
  \country{UK}
}

\author{Ahmed Salem}
\email{t-salemahmed@microsoft.com}
\affiliation{
  \institution{Azure Research}
  \city{Cambridge}
  \country{UK}
}

\author{Shruti Tople}
\email{shruti.tople@microsoft.com}
\affiliation{
  \institution{Azure Research}
  \city{Cambridge}
  \country{UK}
}

\author{Menglin Xia}
\email{mollyxia@microsoft.com}
\affiliation{
  \institution{M365 Research}
  \city{Cambridge}
  \country{UK}
}

\author{Santiago Zanella-B\'eguelin}
\email{santiago@microsoft.com}
\affiliation{
  \institution{Azure Research}
  \city{Cambridge}
  \country{UK}
}

\author{Victor R\"uhle}
\email{virueh@microsoft.com}
\affiliation{
  \institution{M365 Research}
  \city{Cambridge}
  \country{UK}
}

\renewcommand{\shortauthors}{\small L. Wutschitz, B. K\"opf, A. Paverd, S. Rajmohan, A. Salem, S. Tople, M. Xia, S. Zanella-B\'eguelin and V. R\"uhle}
\authorsaddresses{}

\newcommand{\powset}[1]{\mathcal{P}(#1)}

\newcommand{\users}{U}
\newcommand{\allusers}{\mathcal{U}}
\newcommand{\data}{D}
\newcommand{\aclabel}{\ell}

\begin{document}

\begin{abstract}
  
Modern machine learning systems use models trained on ever-growing corpora.
Typically, metadata such as ownership, access control, or licensing information is ignored during training.
Instead, to mitigate privacy risks, we rely on generic techniques such as dataset sanitization and differentially private model training, with inherent privacy/utility trade-offs that hurt model performance.
Moreover, these techniques have limitations in scenarios where sensitive information is shared across multiple participants and fine-grained access control is required.
By ignoring metadata, we therefore miss an opportunity to better address security, privacy, and confidentiality challenges.

In this paper, we take an information flow control perspective to describe machine learning systems, which allows us to leverage metadata such as access control policies and define clear-cut privacy and confidentiality guarantees with interpretable information flows.
Under this perspective, we contrast two different approaches to achieve user-level non-interference: 1) fine-tuning per-user models, and 2) retrieval augmented models that access user-specific datasets at inference time.
We compare these two approaches to a trivially non-interfering zero-shot baseline using a public model and to a baseline that fine-tunes this model on the whole corpus. 
We evaluate trained models on two datasets of scientific articles and demonstrate that retrieval augmented architectures deliver the best utility, scalability, and flexibility while satisfying strict non-interference guarantees.

\end{abstract}


\maketitle

\section{Introduction}

Recent advances in generative machine learning have been enabled by  ever-increasing model sizes and training corpora.
These models show impressive performance on many tasks, most prominently language modelling and image generation.
This has lead to their wide adoption in real world applications.
However, it has also been shown that these models can memorize and subsequently leak information about their training data \citep{carliniextractgpt2,extractingimages}.
To remedy this, models are trained on sanitized datasets \cite{crt,analyze_pii} or with anonymization techniques such as differentially private stochastic gradient descent (DP-SGD) \cite{song, Abadi_2016}.
Originally proposed for privacy-preserving statistical databases, the use of differential privacy is not straightforward in other domains such as natural language, as highlighted by \citet{brownprivacy}.
Moreover, differential privacy is a continuum of guarantees with real-valued parameters $\varepsilon$ and $\delta$.
These parameters are hard to interpret and select, with the debate about what choices are safe for a given application still ongoing. 

\paragraph{Motivation.}
Nowadays, machine learning (ML) models form central components within most web, mobile and desktop applications such as authoring assistants, summarization tools or search engines.
These systems require user authentication, yet often a single model is used to serve all users while metadata such as ownership, access control or licensing information that can be used to make the system more secure and private is discarded during training.
We argue, that when designing for privacy, considering ML components in isolation impedes better solutions that take into account their role as part of a whole system. As a concrete example, we consider an authoring assistant for collaboratively editing a document shared among a group of users. The users aim to write, say, about a project they all work on, without leaking sensitive information (\eg, about a tented project only a subset of them know about). ML models powering this authoring assistant, such as next word prediction or summarization models, should make the best use of data accessible to all collaborating users while stopping information flowing from other more sensitive sources.

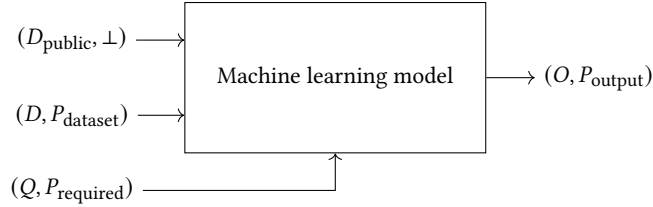
\begin{figure}
  \centering
  \begin{tikzpicture}
[
    roundnode/.style={circle, draw=green!60, fill=green!5, very thick, minimum size=7mm},
    squarednode/.style={rectangle, draw=red!60, fill=red!5, very thick, minimum size=5mm},
]
    \node
        (system)
        [rectangle, draw, minimum height=2cm, minimum width=4cm]
        at (0, 0)
        {Machine learning model};

    \node
        (Dpub)
        at (-3.5, 0.5)
        {$(D_{\textnormal{public}}, \perp)$};
    \node
        (Dpriv)
        at (-3.5, -0.5)
        {$(D, P_{\textnormal{dataset}})$};
    \node
        (Q)
        at(-3.5, -1.5)
        {$(Q,P_{\textnormal{required}})$};
    \node
        (output) at (3.5,0)
        {$(O, P_{\textnormal{output}})$};

    \draw [->] (Dpub) -- (system.west|-Dpub);
    \draw [->] (Dpriv) -- (system.west|-Dpriv);
    \draw [->] (Q) -| (system.south);
    \draw [->] (system.east|-output) -- (output);

\end{tikzpicture}
  \caption{
    Illustration of a machine learning pipeline where inputs and outputs have policies that govern how data can be used.
    This policy could be a set of users that are allowed to access the data.
    In this example, the public pre-training dataset $D_{\textnormal{public}}$ has a trivial policy (indicated by $\bot$), however, the sensitive training dataset $D$ and the query $Q$ have stricter policies.
    In the framework we propose, the model produces an output $O$ with a policy $P_{\textnormal{output}}$ that is compatible with the policies of all inputs.
  }
  \Description{}
  \label{fig:sys}
\end{figure}

\paragraph{Problem.}
\autoref{fig:sys} describes the inputs and output of a system with an ML model as a core component. 
Within this framework, all inputs and outputs of components are pairs of data and metadata which might describe the set of users that are authorized to access the data.
The machine learning model is a component within a pipeline with inputs (training data and queries) and outputs.
The ideal outcome will be to have a clear-cut privacy guarantee for users providing their data without degrading utility for users querying the system.
This leads us to the question: \textit{How can we incorporate metadata into the design of ML models and what privacy/utility guarantees can we provide?}

\paragraph{Threat model}

We assume an adversary that has to authenticate with the system and consequently abides by the policies of the system.
The adversary has full access to the model and its parameters and can query the model with arbitrary inputs.
The adversary also has control over a subset of the training data indicated by the security labels.

\paragraph{Our Position.}
We propose formulating the privacy analysis of ML components as an information-flow control (IFC) problem.
We argue that reasoning about the privacy of ML models from an information flow control perspective allows us to explore richer ML architectures while improving utility.

\paragraph{Contributions.}
Following this perspective, we apply concepts from information flow control to machine learning pipelines and reason about privacy, scalability, and utility properties of end-to-end applications using them.
With this goal, we investigate and compare four privacy-aware approaches to knowledge sharing across users under access control constraints:
\begin{inparaenum}
\item \label{it:zeroshot} a zero-shot baseline using a public pre-trained model,
\item \label{it:dpsgd} a differentially-private baseline that fine-tunes this model on the whole corpus,
\item \label{it:personalized} personalized per-user models fine-tuned only on the data accessible by each user,
\item \label{it:augmented} a retrieval augmented architecture that accesses user-specific datasets at inference time.
\end{inparaenum}
We find that: \cref{it:zeroshot} underperforms compared to the other approaches, \cref{it:dpsgd} can only improve utility while allowing some leakage to happen, 
\cref{it:personalized} have strict privacy guarantees but do not scale to a large number of users, and \cref{it:augmented} enjoys the same strict guarantees, scales well, and provides additional flexibility to changes to policies and data.

\section{Information Flow Control in ML Pipelines}

In this section, we introduce the model and core concepts that we use in the following to reason about information flow in ML pipelines.
In particular, we define non-interference and relate it to the notion of Differential Privacy (DP), adopted from the literature on statistical databases to bound information leakage in ML.

\subsection{Data and Modeling Setup}

\paragraph{Security labels.}
We consider data that is governed by an access control policy.
Consider a set $\data$ of data items and the universe of all users $\allusers$.
Only a subset of the users might have access to a data item $d$; for example, a document might only be accessible to certain employees within a company.
We represent this by assigning a \emph{security label} $\aclabel(d)$ to each data item $d$ to represent the set of users that are authorized to access $d$.
In this setting, security labels can be seen as access control lists (ACLs) for each data item.
\begin{equation}
  \aclabel\colon \data \rightarrow \powset{\allusers} 
\end{equation}

\paragraph{Label updates.}
Security labels may change during the lifetime of the system.
For example, users might be granted access to previously inaccessible data items, or might have their access to existing items revoked.
Furthermore, the set of users accessing the system might change, with users being added or removed.
Formally, this means that security labels also have a time dependence, but for clarity of explanation we always refer to the security label at the current point in time (i.e., assuming label updates propagate instantaneously).

\paragraph{Declassification}
One common type of label update is \emph{declassification} of a data item.
In multi-level security (MLS) systems, this would refer to lowering the security level of the item (\eg, \textsc{CLASSIFIED} $\rightarrow$ \textsc{UNCLASSIFIED}).
In our setting, we use this term to refer to a deliberate and controlled expansion of the set of users who are authorized to access a data item.
In most cases, declassification makes a data item accessible to all users of the system \eg, $\aclabel(d) = \bot$.
Partial declassification might also happen implicitly \eg, when one user shares or sends a document to another user.

\paragraph{Projections.}
Given a dataset $\data = \{d_0 ... d_n\}$ and a set of users $\users \subseteq \allusers$, 
we define the \emph{projection} $\data_\users$ as the subset of $\data$ accessible by \emph{all} users in $\users$,
\begin{equation}
  D_{U} = \{ x \in D: U \subseteq \aclabel(x) \}
\end{equation}
Clearly, when the set of users grows, the data to which they all have access cannot get larger. 
This is captured by the following lemma:
\begin{lemma}
  \label{lm:projection}
  The projection to unions of sets of users is the intersection of the projections to the individual user sets.
  \begin{equation}
  D_{U\cup V}=D_{U}\cap D_{V} \; . \label{eq:projection}
  \end{equation}
\end{lemma}

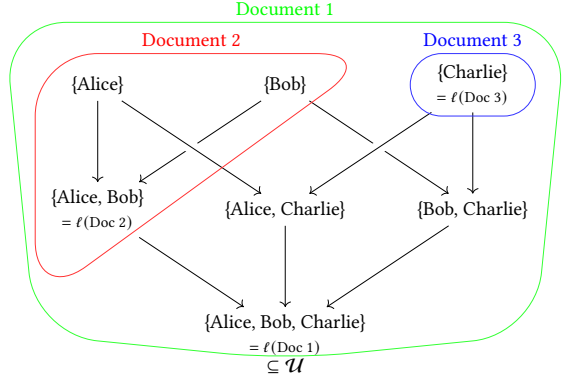
\begin{figure}[ht]
  \centering
  \resizebox{0.5\textwidth}{!}{\begin{tikzpicture}
 \node (a) at (-3,2) { \{Alice\} };
  \node (b) at (0,2) { \{Bob\} };
  \node [align=center] (c) at (3,2) { \{Charlie\} \\ {\footnotesize $=\ell(\text{Doc 3})$} };
  \node [align=center] (ab) at (-3,0) { \{Alice, Bob\} \\ {\footnotesize $=\ell(\text{Doc 2})$} };
  \node (ac) at (0,0) { \{Alice, Charlie\} };
  \node (bc) at (3,0) { \{Bob, Charlie\} };
  \node [align=center] (min) at (0,-2) { \{Alice, Bob, Charlie\} \\ {\footnotesize $=\ell(\text{Doc 1})$} };
  \node (public) at (0,-2.5) {$\subseteq \allusers$};
  \draw [<-] (min) -- (ab);
  \draw [<-] (ab) -- (a);
  \draw [<-] (min) -- (bc);
  \draw [<-] (bc) -- (c);
  \draw [<-] (min) -- (ac);
  \draw [<-] (ac) -- (a);
  \draw [<-] (ac) -- (c);
  \draw [<-] (ab) -- (b);
  \draw [<-] (bc) -- (b);
  \draw[->, preaction={draw=white, -,line width=6pt}] (a) -- (ac);
  \draw[<-, preaction={draw=white, -,line width=6pt}] (ac) -- (c);
  \draw [rounded corners=10mm, draw=red!80](-4,2.5) -- (1.5,2.5) -- (-4,-1.5) -- cycle;
  \draw [rounded corners=10mm, draw=green!80](-4.5,3) -- (-4,-2) -- (0,-2.4) -- (4, -2) -- (4.5, 3) -- cycle;
  \draw [rounded corners=5mm, draw=blue!80](2,2.5) -- (2,1.5) -- (4,1.5) -- (4,2.5) -- cycle;
  \node [anchor=south, text=red] (doc2) at (-1.5, 2.5) {Document 2};
  \node [anchor=south, text=green] (doc1) at (0, 3) {Document 1};
  \node [anchor=south, text=blue] (doc1) at (3, 2.5) {Document 3};
\end{tikzpicture}}
  \caption{
    Confidentiality lattice modeling read access by users.
    Each node represents the data items that are accessible to that set of users.
    Specifically, the nodes in the top row represent the data items accessible to each individual user, and the nodes lower down with sets of users represent data items to which all users in the set have access \eg, shared group chats, meeting transcripts, etc.
    The bottom node in the lattice represents data that is accessible to all users in the system, for which the security label is written as $\bot$.
  }
  \label{fig:lattice}
  \Description{}
\end{figure}

\paragraph{Lattice.}
\autoref{fig:lattice} provides a graphical example of \autoref{lm:projection}.
The power set of $\allusers$ forms a lattice ordered by inclusion.
The security label $\ell$ identifies which nodes in the lattice are authorized to access particular data items.
Sets of users are only authorized to access a data item if each member individually is authorized to access that item, as represented by the arrows.
For example, the user set \emph{\{Alice, Bob\}} has access to \emph{Document~2}, whilst the user set \emph{\{Alice, Charlie\}} does not, even though Alice has access to that particular data item.
Note that since the security labels and/or users may change over time, the projections of datasets onto groups of users and thus the lattice itself might also vary with time.

\paragraph{Machine learning component.}
A machine learning component in an application is responsible for training models and using them to answer users' queries.
Formally, we model an ML component as a randomized function $f$ that takes as input a query $x$, a dataset $D$ and a set of users $\users$.
The output of $f(x, D, \users)$ is a distribution where the randomness comes from stochasticity in the optimization algorithm (typically SGD or a variant thereof).

\subsection{Non-interference}
\label{sec:non-interference}

The tenet of our approach is the property of \emph{non-interference}.
This concept has been widely studied in the security literature \citep{Goguen:1982,Gray:1991,Volpano:1996,Sabelfeld:2009}. We extend it to machine learning pipelines.

\begin{definition}[Probabilistic non-interference]
A machine learning component $f$ satisfies non-interference if for any $\users \subseteq \allusers$, query $x$, pair of datasets $(D, D')$, and set of outputs $S$,
  \begin{equation}
    D_{U} = D'_{U} \implies \Pr{f(x, D, U) \in S} = \Pr{f(x, D', U) \in S} \; . 
    \label{eq:noninterference}
  \end{equation}
\end{definition}

In other words, a function $f$ satisfies non-interference if, for any set of users $U$ and any pair of datasets $D$ or $D'$ with equal projections $D_{U} = D'_{U}$, any data items that are part of either dataset but are \emph{not} in the projection onto $U$ (\ie, $D \setminus D_{U}$ or $D' \setminus D'_{U}$) cannot affect the output of function $f$.

Non-interference naturally captures the confidentiality property we desire.
By definition, the set of users $U$ (collectively) should not have access to any data items in $D$ outside $D_{U}$.
A function satisfying non-interference thus ensures that these inaccessible data items do not affect the output.
For example, if a set of users $U$ only has access to a subset of $D$ (\ie, $D_{U} \subset D$), we could construct a hypothetical dataset $D'$ containing only the data items to which $U$ has access (\ie, $D_{U} = D'_{U} = D'$).
By the definition of non-interference, the output of the function would not differ irrespective of whether $D$ or $D'$ were used.

In our example of a group of users using an authoring assistant to write a document, the  assistant must satisfy non-interference with respect to the data items it uses (\eg, other documents).
Following \autoref{lm:projection}, the group of users should only have access to documents that are accessible to all its members. 
By satisfying non-interference, the authoring assistant would only use these documents to produce its outputs, \ie, it would guarantee \emph{perfect secrecy}.
Even if users attempt to run membership or attribute inference attacks against the authoring assistant, they would only learn information about documents to which they already have access.

\subsection{Source Attribution}
\label{sec:source-attribution}

In addition to non-interference, another desirable property is the ability to attribute the output of a function to a subset of the input, \ie, \emph{source attribution}.

\begin{definition}[Source attribution]
  Let $f$ be an ML component. 
  When $f$ answers a query $x$ for a group of users $\users$ (\ie, at security label $\users$) given a dataset $D$, it uses some subset of data items in $D$.
  We call any such possible subset a \emph{source} of $f(x, D, \users)$.
  Formally, a source of $f(x, D, \users)$ is any $D' \subseteq D$ for which:
  \begin{equation}
    f(x, D', U) = f(x, D, U) \; .
  \end{equation}
\end{definition}

A source of $f(x, D, \users)$ is any subset of data items in $D$ that could have been used to produce the output.
Note that $D$ itself is always a source, and so is $D_\users$ when $f$ satisfies non-interference. 
However, $D_\users$ would typically be an overapproximation and the set of data items determining the model output could be smaller.
In \Cref{sec:architectures}, we discuss different ways to reason about source attribution (\eg, based on the size of the sources set) for different ML architectures.

In general, surfacing source information along with an output enables users to understand which data items were used by an ML model when processing a query, thereby allowing better interpretability.
This is desirable for many reasons.
For example, before sending an email that was generated with the help of an authoring assistant, the sender may check which sources the assistant used.
From a confidentiality perspective, the recipient may not have access to some of these sources, and thus the sender must decide whether it is appropriate to share (\ie, declassify) the output generated by the assistant.
On the other hand, even if the recipient has access to all the sources, the sender may want to check that the sources only consist of trustworthy documents, to mitigate potential integrity issues.

\subsection{Scalability}
For practical deployment, it is important that the design of the ML component is scalable to changes in the lattice structure such as either addition/deletion of users or label updates to the existing documents. Examples of such modifications may happen when an employee 1) leaves a company or 2) transfers to a different organization or 3) moves to a different project within the same organization. Any of the these scenarios may introduce different changes to the lattice structure and the ML component should scale appropriately in a computationally efficient manner. The need for computationally expensive operations (such as retraining entire models) to support every single change may push the application to delay updates to the ML component and perform them in an infrequent manner. However, failure to adapt quickly may result in privacy concerns by leaking sensitive information when access is revoked, or degrade utility by not using relevant documents when access is granted.

\subsection{Differential Privacy in the Context of Information Flow}
\label{sec:sec2dp}

Differential privacy (DP) is the state-of-the-art technique for privacy preserving data processing.
DP achieves this by bounding the contribution of each user and adding calibrated noise.
The privacy risk of the resulting model is captured by continuous parameters $\varepsilon$ and $\delta$.

Let $D \sim D'$ when $D$ and $D'$ differ only in the data of a single user. 
A randomized function $f$ satisfies user-level $(\varepsilon,\delta)$-DP if for every pair of datasets ($D$, $D'$) and all sets of outputs $S$,

\begin{equation}
  D \sim D' \implies \Pr{f(D) \in S} \leq e^{\varepsilon} \Pr{f(D') \in S} + \delta \; .
\end{equation}

Intuitively, the smaller $\varepsilon$ (and $\delta$) the better the privacy protection.
The additive term $\delta$ is usually chosen as $\nicefrac{1}{n\log n}$ where $|\allusers| \leq n$.
Even though theoretically only $\varepsilon \leq 1$ could be considered to provide any meaningful privacy protection, in practice values up to 10 have shown to mitigate concrete privacy attacks \citep{lira}.
In the end, what values of $\varepsilon$ mitigate privacy risks depends on the application and data distribution.
Once this is agreed, training with differential privacy can be considered a \emph{declassification} and the output label may be more permissive than the inputs.

\paragraph{Degradation of privacy for groups of users.}
However, protecting confidentiality not only requires picking a suitable $\varepsilon$, but also ensuring that the data distribution is compatible with the flavor of DP guaranteed by the training algorithm.\footnote{
  There has been much debate whether DP is meant to only protect the presence (or absence) of individual data contributors or the disclosure of sensitive information \citep[Appendix A]{causal_dp_sok}.
  In this paper, we are interested in protecting sensitive information and therefore require independence of users' data contributions.
}
Under user-level differential privacy, data items contributed by groups of users see their privacy degraded to a group privacy guarantee~{\cite[Theorem 2.2]{dpbook}}.
This can be problematic if DP is used to protect sensitive data rather than the mere presence or absence of users in the dataset.
While document- and sentence-level deduplication can alleviate this problem, deduplication techniques are imperfect and leave correlations between the data of different users \citep{Kandpal:2022}.

\paragraph{Relationship to non-interference}

One notable difference between DP and non-interference is that DP assumes that there is a single user that contributes a data item $d$ (typically the author).
Whereas in information flow control, the security label $\aclabel(d)$ determines access.
Assuming each user has access to the data they contribute, \emph{non-interference} and $(\varepsilon=0, \delta=0)$-DP are equivalent \citep{causaldp}, making \emph{non-interference} a stricter property.

\section{Comparing Different ML Pipelines}
\label{sec:architectures}

Having laid the groundwork of introducing information flow control concepts and expressing desirable properties of machine learning pipelines, we now turn to the investigation of whether existing ML approaches satisfy these properties and identify shortcomings, if any.
Using our example of collaborative document authoring assistant, we compare three different approaches (i) a single model fine-tuned on the whole sensitive dataset, (ii) multiple models each fine-tuned on a node in the information flow lattice, and (iii) retrieval augmented architectures that retrieve data at inference time.

\subsection{Fine-tuning a Single Global Model}

In this approach, we assume that a central model is fine-tuned on a sensitive dataset.
It has been shown that large image and language models memorize training data and produce near identical copies \citep{extractingimages,carliniextractgpt2}.
In the language of information flow control, only users with access to the whole dataset (if there are any) could safely access the model.
Thus, making the model available to more users amounts to a declassification.

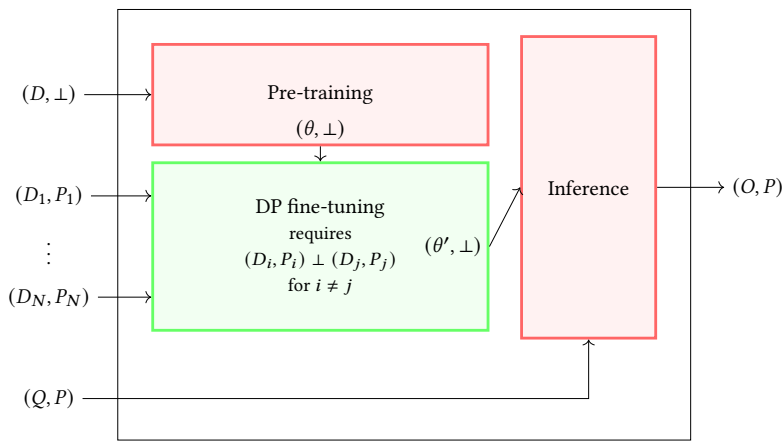
\begin{figure*}
  \centering
  \resizebox{0.7\textwidth}{!}{\begin{tikzpicture}
[
    roundnode/.style={circle, draw=green!60, fill=green!5, very thick, minimum size=7mm},
    squarednode/.style={rectangle, draw=red!60, fill=red!5, very thick, minimum size=5mm},
]
    \node
        (Dpub)
        {$(D, \perp)$};
    \node
        (D1)
            [below=of Dpub]
            {$(D_1, P_1)$};
    \node
        (DN)
        [below=of D1]
        {$(D_N, P_N)$};
    \node
        [squarednode, minimum width=5cm, minimum height=1.5cm]
        (train)
        [right=of Dpub]
        {Pre-training};
    \node
        [anchor=south]
        (theta)
        at (train.south)
        {$(\theta, \perp)$};
    \node
        (dots1)
        at ($(D1)!0.5!(DN)$)
        {$\vdots$};
    \node
        [squarednode, minimum height=2.5cm, minimum width=5cm, align=center, fill=green!5, draw=green!60]
        (finetuning)
        at (dots1-|train)
        {DP fine-tuning \\ {\small requires} \\ {\small $(D_i, P_i) \perp (D_j, P_j)$} \\ {\small for $i\neq j$}};
    \node
        [anchor=east]
        (theta2)
        at (finetuning.east)
        {$(\theta^{\prime}, \perp)$};

    \node
        (Q)
        [below=1cm of DN]
        {$(Q,P)$};

    \node
        [squarednode,xshift=4cm, minimum height=4.5cm, minimum width=2cm]
        (inference)
        at ($(train.north)!0.5!(finetuning.south)$)
        {Inference};
    \node
        (output)
        [right=of inference]
        {$(O, P)$};
    
    \draw[->] (Dpub.east) -- (train.west);
    \draw[->] (D1.east) -- (D1.east-|finetuning.west);
    \draw[->] (DN.east) -- (DN.east-|finetuning.west);
    \draw[->] (train.south) -- (finetuning.north);
    \draw[->] (Q.east) -| (inference.south);
    \draw[->] (finetuning.east) -- (inference.west);
    \draw[->] (inference.east) -- (output.west);

    \draw[draw=black]
        ($(train.north west)+(-0.5,0.5)$) rectangle ($(inference.south east)+(0.5,-1.5)$);

\end{tikzpicture}}
  \caption{
    System diagram of a DP-fine-tuning pipeline.
    Public data $(D, \perp)$ is used for pre-training which produces public model weights $(\theta, \perp)$.
    The model is then fine-tuned on private data where the input data has to be pairwise independent.
    This is a requirement for DP and may be difficult to guarantee in practical scenarios.
    DP fine-tuning is a declassification step making the resulting model weights $\theta^{\prime}$ public.
    This trivially satisfies the policy requirement $P$ from the input query $Q$.
  }
  \label{fig:sys_dp}
  \Description{}
\end{figure*}

\autoref{fig:sys_dp} illustrates a possible instantiation of a pipeline with a central model and declassification.
The model is trained using DP-SGD with a DP guarantee that the operator has deemed sufficient for a declassification.
In addition, for this approach to work, each data contributor has to be independent.
The model is trained using a two-stage transfer learning setup that has been shown to significantly improve utility when training with DP-SGD \citep{dpfinetuning,dpneedsbetterfeatures,dpscale,lilargelanguagemodelsdp}.
First, the model is trained without any privacy guarantees on a public data corpus $(D, \perp)$ producing weights $(\theta, \perp)$.
Subsequently, the model is fine-tuned with DP, producing updated weights $\theta^{\prime}$ which are then declassified ($\aclabel(\theta^\prime) = \bot$).
Finally, the inference output is computed from the weights and the query.
The output inherits the policy from the query since the query may leak information to the output.

Revisiting the example of the collaborative authoring assistant, we emphasize that the requirement of independent data contributors in DP is often an unrealistically restrictive assumption.
\citet{brownprivacy} pointed out that for datasets where people communicate and collaborate there may be large groups unknown to the data processor which can severely degrade privacy guarantees.
Another shortcoming of this approach is the limited ability to handle label and data updates.
If new data becomes available or labels change, and the privacy budget has been exhausted, prior data cannot be used to update the model.
Updated models need to either use additional privacy budget or be trained on a disjoint set of users.
It is possible to release models with a smaller $\varepsilon$ than is needed for declassification to save privacy budget for future releases, however, this adds significant complexity to the approach and further harms utility.

\subsection{Fine-tuning Personalized Models}

A second approach for preserving users' privacy is to train personalized models for each node of the lattice. 
This approach involves fine-tuning a public model on each of the users' private data.
Concretely, each set of users $U$ has their own fine-tuned model $g_{\theta(D_U)}$ that is trained on their data $D_U$.
This approach achieves non-interference as each node's model only sees data the users own or have access to during training.
This can be seen by writing the training and inference process as
\begin{equation}
  f(x, D, U) = g_{\theta(D_U)}(x) \; .
\end{equation}
Note that $f(x, D, U) = f(x, D_U, U)$ which implies non-interference.
 
While this approach has significantly better utility with privacy by design compared to differential privacy, it is not practical for large groups of users.
Training and maintaining a separate model for every node requires a significant amount of computational and storage resources, making it difficult to scale.

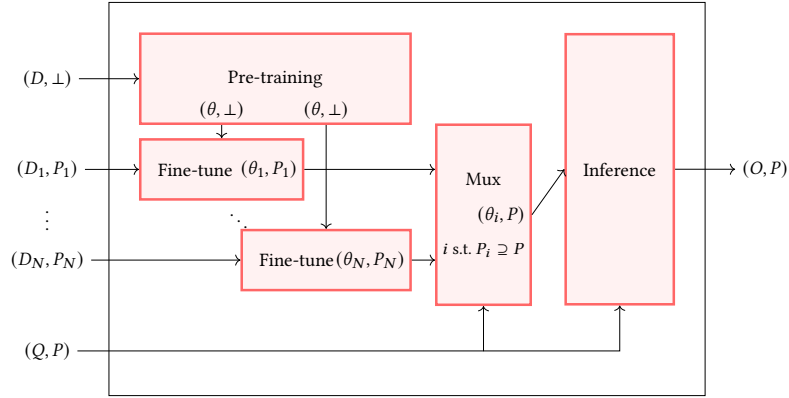
\begin{figure*}
  \centering
  \resizebox{0.7\textwidth}{!}{\begin{tikzpicture}
[
    squarednode/.style={rectangle, draw=red!60, fill=red!5, very thick},
]
    \node
        (Dpub)
        {$(D, \perp)$};
    \node
        (D1)
            [below=of Dpub]
            {$(D_1, P_1)$};
    \node
        (DN)
        [below=of D1]
        {$(D_N, P_N)$};
    \node
        [squarednode, minimum width=4.5cm, minimum height=1.5cm]
        (train)
        [right=of Dpub]
        {Pre-training};
   \node
        (dots1)
        at ($(D1)!0.5!(DN)$)
        {$\vdots$};
    \node
        [squarednode, xshift=2cm, minimum height=3cm, anchor=east, align=center]
        (mux)
        at (dots1-|train.east)
        {Mux \\ \\ \\ {\small $i$ s.t. $P_i \supseteq P$}};
    \node
        [align=left, squarednode, anchor=west, minimum height=1cm, text width=2.5cm]
        (M1)
        at (train.west|-D1)
        {\; Fine-tune};
    \node
        [anchor=east]
        (theta_i)
        at (mux.east)
        {$(\theta_i, P)$};
    \node
        [anchor=east]
        (theta1)
        at (M1.east)
        {$(\theta_1, P_1)$};
    \node
        [anchor=west]
        (Mdots)
        at (M1.south|-dots1)
        {$\ddots$};
    \node
        [align=left, squarednode, anchor=east, minimum height=1cm, text width=2.6cm]
        (MN)
        at (train.east|-DN)
        {\; Fine-tune};
    \node
        [anchor=south]
        (theta)
        at (train.south-|MN)
        {$(\theta, \perp)$};
    \node
        [anchor=south]
        (theta)
        at (train.south-|M1)
        {$(\theta, \perp)$};
    \node
        [anchor=east]
        (thetaN)
        at (MN.east)
        {$(\theta_N, P_N)$};
    \node
        (Q)
        [below=1cm of DN]
        {$(Q,P)$};
    \node
        [squarednode,xshift=4cm, minimum height=4.5cm, minimum width=1.8cm]
        (inference)
        at ($(train.north)!0.5!(mux.south)$)
        {Inference};
    \node
        (output)
        [right=of inference]
        {$(O, P)$};
    
    \draw[->] (Dpub.east) -- (train.west);
    \draw[->] (D1.east) -- (D1.east-|M1.west);
    \draw[->] (DN.east) -- (DN.east-|MN.west);
    \draw[->] (train.south-|M1.north) -- (M1.north);
    \draw[->] (train.south-|MN.north) -- (MN.north);
    \draw[->] (Q.east) -| (mux.south);
    \draw[->] (Q.east-|mux.south) -| (inference.south);
    \draw[->] (M1.east) -- (mux.west|-M1.east);
    \draw[->] (MN.east) -- (mux.west|-MN.east);
    \draw[->] (mux.east) -- (inference.west);
    \draw[->] (inference.east) -- (output.west);

    \draw[draw=black]
        ($(train.north west)+(-0.5,0.5)$) rectangle ($(inference.south east)+(0.5,-1.5)$);

\end{tikzpicture}}
  \caption{
    System diagram of a pipeline with personalized models.
    Public data $(D, \perp)$ is used for pre-training which produces public model weights $(\theta, \perp)$.
    Subsequently, a new model is fine-tuned for each non-empty node in the lattice $((\theta_1, P_1), \dots, (\theta_N, P_N))$.
    A model selection stage (Mux) then ensures that only model weights are used that are compatible with the policy $P$.
  }
  \label{fig:sys_pers}
  \Description{}
\end{figure*}

\subsection{Inference-time Policy Enforcement via Retrieval Augmentation}

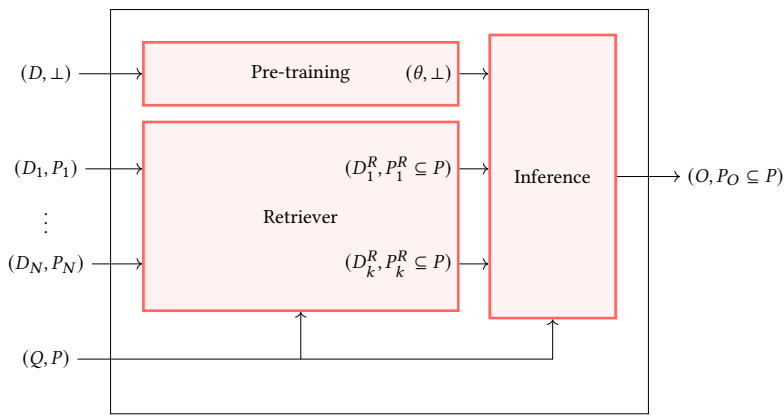
\begin{figure*}[h]
  \centering
  \resizebox{0.7\textwidth}{!}{\begin{tikzpicture}
[
    roundnode/.style={circle, draw=green!60, fill=green!5, very thick, minimum size=7mm},
    squarednode/.style={rectangle, draw=red!60, fill=red!5, very thick, minimum size=5mm},
]
    \node
        (Dpub)
        {$(D, \perp)$};
    \node
        (D1)
            [below=of Dpub]
            {$(D_1, P_1)$};
    \node
        (DN)
        [below=of D1]
        {$(D_N, P_N)$};
    \node
        [squarednode, minimum width=5cm, minimum height=1cm]
        (train)
        [right=of Dpub]
        {Pre-training};
    \node
        [anchor=east]
        (theta)
        at (train.east)
        {$(\theta, \perp)$};
    \node
        (dots1)
        at ($(D1)!0.5!(DN)$)
        {$\vdots$};
    \node
        [squarednode, minimum height=3cm, minimum width=5cm]
        (retriever)
        at (dots1-|train)
        {Retriever};
    \node
        [anchor=east]
        (DR1)
        at (retriever.east|-D1)
        {$(D^R_1, P^R_1\subseteq P)$};
    \node
        [anchor=east]
        (DRk)
        at (retriever.east|-DN)
        {$(D^R_k, P^R_k\subseteq P)$};

    \node
        (Q)
        [below=1cm of DN]
        {$(Q,P)$};

    \node
        [squarednode,xshift=4cm, minimum height=4.5cm, minimum width=2cm]
        (inference)
        at ($(train.north)!0.5!(retriever.south)$)
        {Inference};
    \node
        (output)
        [right=of inference]
        {$(O, P_O \subseteq P)$};
    
    \draw[->] (Dpub.east) -- (train.west);
    \draw[->] (D1.east) -- (D1.east-|retriever.west);
    \draw[->] (DN.east) -- (DN.east-|retriever.west);
    \draw[->] (Q.east) -| (retriever.south);
    \draw[->] (Q.east-|retriever.south) -| (inference.south);
    \draw[->] (theta-|train.east) -- (theta-|inference.west);
    \draw[->] (DR1-|retriever.east) -- (DR1-|inference.west);
    \draw[->] (DRk-|retriever.east) -- (DRk-|inference.west);
    \draw[->] (inference.east) -- (output.west);

    \draw[draw=black]
        ($(train.north west)+(-0.5,0.5)$) rectangle ($(inference.south east)+(0.5,-1.5)$);

\end{tikzpicture}}
  \caption{
    System diagram of a retrieval augmented pipeline.
    Public data $(D, \perp)$ is used for pre-training which produces public model weights $(\theta, \perp)$.
    Based on the query $Q$, $k$ relevant input data samples are retrieved that satisfy the policy requirement $P$.
    The inference step consumes the additional retrieved samples and produces the output $O$ with policy $P_O = \bigcap P^R_i \subseteq P$.
    No declassification step is needed.
  }
  \label{fig:sys_ra}
  \Description{}
\end{figure*}

Retrieval augmented models have gained increasing popularity in recent literature due to their ability to improve the performance of machine learning models \cite{lewis2020retrieval, pmlr-v162-borgeaud22a}.
Retrieval augmentation enhances a machine learning model by adding a memory component that retrieves and integrates external information into the model.
In addition to the performance benefits, retrieval augmented models offer an alternative solution to preserving users' privacy by enforcing access control policies during inference (\autoref{fig:sys_ra}). 
In this pipeline, the base machine learning model is trained on public data to which all users have access, and private data is only retrieved at inference time, while adhering to the security labels.
This ensures that the information surfaced by the model is only what the user is authorized to see.
Therefore, this approach provides robust privacy guarantees by enforcing access control policies.

There are various implementation techniques for retrieval augmentation, which can be broadly classified into four categories: prompting methods, decoding methods, cross-attention methods, and vector fusion methods.
Prompting methods augment a pre-trained model by incorporating retrieved data into the model input through prompt design or prompt tuning \cite{lewis2020retrieval, shi2023replug}.
Decoding methods integrate retrieved information at the model decoding phase by modifying the output distribution of the original model \cite{Khandelwal2020Generalization, liu2022knowledge}.
Cross attention and vector fusion techniques merge the vector representations of the retrieved information with the original model through methods such as simple vector concatenation, weighted summation, or more sophisticated techniques such as cross-attention \cite{fan-etal-2021-augmenting, izacard-grave-2021-leveraging, pmlr-v162-borgeaud22a}.
For an overview of implementations in previous studies and a comparison of these techniques, refer to \Cref{sec:retrieval_augmentation}.

Despite differences in implementation, all retrieval augmentation techniques can be represented by
\begin{equation}
  f(x, D, U) = g_{\theta}(x, D_U) \; ,
\end{equation}
where $g_{\theta}$ is a pretrained model, typically trained on public data without access restrictions.
Again, it is easy to see that $f(x, D, U) = f(x, D_U, U)$ which implies non-interference.

\subsection{Qualitative Comparison of Desired Properties}

\begin{table*}[h]
\caption{Comparison of three ML pipelines with respect to the desirable properties}
\renewcommand{\arraystretch}{1.5}
\begin{tabular}{@{}lccc@{}}
\toprule
\textbf{Approach}               & \makecell{\textbf{fine-tuned global model} \\ (w/ DP)}            & \makecell{\textbf{fine-tuned}\\ \textbf{personalized model}}     & \makecell{\textbf{retrieval augmented}\\ \textbf{model}}  \\ \midrule
Non-interference                & \makecell{No \\ {\footnotesize (Except $\varepsilon=\delta=0$)}}            & \makecell{Yes}                                                   & Yes       \\
Utility                         & \makecell{Degrades \\ {\footnotesize (depends on $\varepsilon$)}} & \makecell{High}                                                  &  High                            \\ 
Handling updates                & \makecell{No \\ {\footnotesize (Requires retraining if privacy budget allows)}}            & \makecell{No \\ {\footnotesize (Requires retraining)}}           &  Yes         \\
Scalable                        & \makecell{Yes \\ {\footnotesize (Single model)}}                        & \makecell{No \\ {\footnotesize (Model per node in the lattice)}} & \makecell{Yes \\ {\footnotesize (Single model)}} \\
Source Attribution              & \makecell{No \\ {\footnotesize (Entire Dataset)}}                 & Within user / node level &  Within retrieved documents \\
 \bottomrule 
\end{tabular}
\label{tab:capabilities}
\end{table*}

\autoref{tab:capabilities} summarizes the capabilities of the three ML pipelines across the different properties.
In the following section, we discuss each capability in more detail.

\paragraph{Non-interference.}
Both the personalized model and inference time policy enforcing architectures provide non-interference by design while maintaining the utility of the model.
The approach of fine-tuning a global model with differential privacy satisfies non-interference only at $\varepsilon=0$.
However, this can only be attained by ignoring sensitive data during training altogether.

\paragraph{Scalability.} 
As the number of users in the dataset grows, the number of nodes in the associated lattice grows exponentially.
This poses a significant scalability problem for the personalized model approach since we need to train one model per node in the lattice.
The central model and the inference time enforcement approach only require a single model making these approaches better suited for large scale deployments.

\paragraph{Handling updates.}
Label updates occur frequently in many practical scenarios.
The personalized model approach can only incorporate changes in the lattices by retraining models.
This is an expensive and time-consuming task especially considering the scale involved as discussed in the previous paragraph.
However, the central model with DP approach is even more rigid.
Due to the nature of the DP guarantees an update of models requires to compose the privacy budget with previous releases.
Once the maximum privacy budget is reached no further updates are possible.
Only the inference time enforced policy approach allows a dynamic update of both lattice and dataset.

\paragraph{Attribution.} 
The inference time approach supports a fine-grained source attribution restricted to $K$ documents as compared to the other approaches.
The personalized model supports attribution at node level \ie, all the documents that are accessible by the users in a node, while a global model can only attribute an output with certainty to the entire dataset.

\paragraph{Summary}
Based on the comparison in \autoref{tab:capabilities}, we conclude that models with policy enforced during inference time are most suited to be used as ML components in end-to-end applications that contain user metadata. These models satisfy the desired properties of privacy, utility and scalability as compared to other alternative approaches. Approaches such as fine-tuning personalized models or a single model with DP are computationally inefficient as they enforce the required policy for privacy during training time. Such training time solutions may be an option for a fixed or static lattice where the nodes or security labels do not change during the lifecycle of the application. However, this is often not the case as it limits the scalability and utility of the application.
\section{Experimental Evaluation}
\label{sec:eval}

Following the theoretical comparison of the three machine learning pipelines, we now turn to the experimental evaluation.
We emphasize that the goal of this section is to compare the approaches as fairly as possible with each other and not to achieve state-of-the-art results with a specialized pipeline.
To this end, we use the same GPT-3 base model (version \textit{davinci}) for all approaches \cite{gpt3}.
For the central model approach, we fine-tune a non-private baseline of GPT-3 on the whole sensitive dataset, since the fine-tuning API of GPT-3 does not offer DP-SGD.
This is an unrealistic setting as the risk of leakage is unmitigated.
Yes, it serves as a utility benchmark for all DP trained models since the utility-privacy trade-off in this setting yields a model that is at most as good as the non-private fine-tuned model.
For retrieval augmented models, we use GPT-3 in a few shot setting and augment the prompt with retrieved sentences from the dataset. 

We revisit the earlier introduced example of a collaborative document authoring scenario.
To simulate this, we require datasets with clear authorship information and complex overlapping communication patterns within the dataset.

\subsection{Dataset}

To evaluate our retrieval augmented models, we need datasets that has both the notions of users and of security labels (\eg, access control information) associated with each data item.
However, this type of information is typically not available in commonly-used ML datasets. 
Therefore, we take two existing datasets and generate the necessary security labels from existing metadata.

\begin{description}
	\item[Elsevier]: The Elsevier dataset~\cite{elsevier} contains 40k (40,091) journal articles (\ie, text documents) from a range of scientific disciplines. Each article consists of an abstract, the main body of text, and authors' information. We exclude articles with incomplete authorship and abstract information, resulting in a dataset of 39,708 data points.
	We treat document authors as users of the system, and assume that each user has access to all the documents they have authored (including co-authorship).
	This is a reasonable assumption for a collaborative document editing scenario.

	\item[Arxiv]: The Arxiv dataset\footnote{https://www.kaggle.com/datasets/Cornell-University/arxiv} contains both metadata and PDF files of open access scientific articles. 
	Similar to the Elsevier dataset, we processed the raw Arxiv PDFs and metadata into abstract, main body text, and authors' information.
	We filtered the dataset by retaining articles created after October 2021. This step was undertaken because the GPT-3 model might have been trained using Arxiv data up until September 2021.
\end{description}

The lattice structure introduced in \autoref{fig:lattice} maps readily to the co-authorship setting in both datasets.
The nodes in the top row of the lattice represents individual authors, and all possible co-author combinations follow below.
Most of the authors have one publication, making these cases less relevant as there is not enough personalized data.
However, there are approximately 1,926 Elsevier articles and 50,019 Arxiv articles where authors have other publications. These data points are then used as our test sets.
We evaluate the models on the language modelling task. We use the body texts of the articles as our training data or retrieval source and evaluate the language model perplexity on the abstracts of the test sets. 
The task of predicting the abstract mimics an authoring task with different levels of access to external information, and our aim is not to simulate a summarization task.

\subsection{Experiments and Results}

To measure the utility of the retrieval augmented models, we first specify the set of accessible documents for predicting each abstract, i.e. the projection of a security label $D_U$.
We define the following variants of the security labels:
\begin{description}
	\item[Access to authors' articles:] The model has access to a set of articles that are accessible to all authors of the current article for the prediction of each abstract.
	\item[Access to authors' other articles:] The model has access to the authors' \textit{previously} written articles, excluding the current article, for the prediction of each abstract. This describes the situation where the author writes about a related topic, but no documents exist yet on this particular topic.
	\item[Access to current article:] The model has access to the corresponding article body of the current abstract. While this setting does not impose any additional access restrictions compared to author's articles, it simulates having the most relevant information for predicting the abstract.
	\item[Access to random:] The model only has access to articles of a randomly selected author, who is not an author of the current article. This mimics that this random author tries to write the article, without having access to any relevant information for predicting the abstract.
\end{description}

We then retrieve augment a pre-trained GPT-3 model by retrieving texts from $D_U$ and evaluate the model performance on the test set. To implement retrieval augmentation, we use the initial tokens of the abstract as the prompt for retrieval. We compare the results of using different proportions (10\%, 25\%, 50\%) of the abstract as the prompt.
For information retrieval, we leverage a pre-trained dual-encoder model \texttt{msmarco-distilbert-base-v4} from the \texttt{sentence-bert} library\footnote{https://www.sbert.net/docs/pretrained-models/msmarco-v3.html} to retrieve the top $K$ most relevant text chunks to the prompt from the respective $D_U$. 

Subsequently, we take the prompting approach for retrieval augmentation and augment the prompt by prepending the retrieved information as additional text tokens to the original prompt. 
We condition the GPT-3 model on the augmented prompt to predict the remaining portion of the abstract. Performance of the model is evaluated in terms of perplexity on the predicted portion of the abstract.
We experimented with different $K$ values and report the best results in this paper and additional results are included in the Appendix.

We compare the performance of the retrieval augmented models to the following baselines:
\begin{description}
    \item[Zero-shot baseline:] We use GPT-3 in a zero-shot next word prediction task to generate an abstract and compute its perplexity. This serves as a non-private baseline to benchmark the utility of a public model without utilizing private data.
    \item[Full fine-tuning baseline:] We fine-tune GPT-3 on all article bodies. This acts as a baseline for DP models. We fine-tune GPT-3 with LoRA, a parameter efficient approach, \cite{hu2022lora} on the next sentence prediction task using the entire training set.
	We perform a hyperparameter search and present the best fine-tuning result in the paper. Details of the hyperparameter search are included in the Appendix.
	\item[Personalized baseline:] Due to constraints in resources, we establish a personalized baseline on a randomly selected subset of 50 abstracts. For each abstract, we develop a personalized model by fine-tuning GPT-3 on the documents that are accessible to the authors of the corresponding article.
\end{description}
To ensure a fair comparison, the baseline models are conditioned on the same proportions of the abstracts as the retrieval augmented models and evaluated on the predicted portion of the abstracts.

\begin{table*}[t]
	\caption{Perplexity scores of retrieval augmented models compared to the baselines, with lower scores indicating improved performance. }
    \begin{tabular}{lcccccc}
    \toprule
    \textbf{Dataset} & & \textbf{Elsevier} & & & \textbf{Arxiv} & \\
    \textbf{Prompt percentage} & \textbf{10\%} & \textbf{25\%} & \textbf{50\%} & \textbf{10\%} & \textbf{25\%} & \textbf{50\%}  \\ \midrule
    \textit{Baseline}: Zero-shot baseline & 10.9 & 10.5 & 10.4 & 12.6 & 12.1 & 11.8 \\
    \textit{Baseline}: Full fine-tune baseline & 10.6 & 10.3 & 10.2 & 13.9 & 13.5 & 13.0 \\
    \textit{RA}: Access to authors' articles & 8.60 & 7.79 & 7.81 & 7.82 & 7.49 & 7.40 \\
    \textit{RA}: Access to authors' other articles & 9.71 & 9.50 & 9.50 & 10.6 & 10.4 & 10.3 \\
    \textit{RA}: Access to current article & 7.40 & 7.27 & 7.39 & 7.00 & 6.84 & 6.84 \\
    \textit{RA}: Access to random & 11.2 & 10.7 & 10.5 & 13.2 & 12.6 & 12.2 \\ 
    \bottomrule
\end{tabular}

	\label{tab:result}
\end{table*}

\autoref{tab:result} shows the performance of the retrieval augmented models compared to the baselines on the full test sets. 
In our study, the GPT-3 zero-shot baseline already exhibited good performance in predicting abstracts on both datasets.
Fine-tuning the GPT-3 model on all article bodies only improves the zero-shot baseline by approximately 2.5\% on the Elsevier dataset despite hyperparameter search. And on the Arxiv dataset, the fine-tuned model performs worse than the zero-shot baseline.

Nonetheless, retrieval augmented models that leverage private data show clear improvements over the baseline models.
As expected, the extent to which retrieval augmentation can improve model performance is correlated to the relevance of information present in the retrieval source.
Specifically, leveraging the current article for retrieval augmentation yields the best results, followed by leveraging articles that all authors have access to. This is in fact a superset of the current article. 
Even when the current article is not accessible, retrieving information from other articles authored by the same authors still leads to improved model performance compared to the baselines. Unsurprisingly, retrieving information from a randomly selected author results in decreased performance.
Furthermore, retrieval augmentation shows a pronounced improvement when the models are conditioned on a smaller fraction of the abstract. For more detailed analysis on model utility, see Appendix \ref{sec:utility}. 

\autoref{tab:personalization} presents the perplexity scores of the personalized baseline and the retrieval augmented model that accesses the same data on the 50 data samples. To provide a better comparison, the results of the other two baselines on these samples are also included.
The personalized model sets a stronger baseline compared to the zero-shot and full fine-tuning baselines. Nonetheless, a retrieval augmented model, which utilizes the same data, can attain results that are on par with or surpass the performance of the personalized model.

\begin{table*}[t]
	\caption{Personalization baseline on 50 randomly selected data points}
    \begin{tabular}{lcccccc}
    \toprule
    \textbf{Dataset} & & \textbf{Elsevier} & & & \textbf{Arxiv} & \\
    \textbf{Prompt percentage} & \textbf{10\%} & \textbf{25\%} & \textbf{50\%} & \textbf{10\%} & \textbf{25\%} & \textbf{50\%}  \\ \midrule
    \textit{Baseline:} Zero-shot baseline & 10.6 & 10.2 & 10.0 & 12.4 & 11.8 & 11.5 \\
    \textit{Baseline:} Full fine-tune baseline & 10.4 & 9.97 & 9.91 & 14.1 & 13.5 & 13.0 \\
    \textit{Baseline:} Personalized baseline & 8.50 & 8.39 & 8.60  & 10.6 & 10.3 & 10.3 \\
    \textit{RA:} Access to authors' articles & 8.42 & 7.77 & 8.00 & 8.45 & 8.00 & 7.55 \\ \bottomrule
\end{tabular}
	\label{tab:personalization}
\end{table*}

\section{Discussion}

\subsection{Related Work}

\paragraph{Critiques of training language models with DP}

Recently, there haven been several papers studying limitations of applying DP to transfer learning as a privacy mitigation.
\citet{brownprivacy} have investigated shortcomings of existing privacy preserving techniques for training language models.
They concluded that many techniques including data sanitization as well as differential privacy have limitations for complex datasets such as natural language. 
Furthermore, \citet{privatepretraining} have reviewed the popular DP fine-tuning approach concluding that applying DP only during fine-tuning may not be sufficient and more work is required to ensure pre-training datasets are private.

\paragraph{Privacy for machine learning models}
\citet{tiwari2023information} consider applying information flow control to machine learning models during training via a mixture of experts architecture.
\citet{huang2023privacy} have studied the privacy implications of retrieval-based language models but mainly focused on $k$NN-LM architectures.
\citet{foundationsecrecy} have studied large language model's ability to utilize in context learning (ICL) as an alternative to federate learning (FL).
Their approach is competitive with FL baselines while providing stronger privacy guarantees.

\paragraph{Relation to current privacy attacks.}
The integration of information flow in machine learning architectures results in a privacy by design model.
This renders traditional privacy attacks, such as membership inference \cite{SSCS17,SZHBFB19,NSH18}, model inversion \cite{JG18,WFJN16}, attribute inference \cite{ZCBSZ22,FJR15}, property inference \cite{GWYGB18,mahloujifar2022property}, and data reconstruction \cite{Carlini:2019,Carlini:2021c}, irrelevant.
Since with this model, users only receive output from their own documents, gaining no additional information beyond what they can obtain by directly accessing the documents.

\subsection{Future work}

While the evidence is compelling to use retrieval augmented models to implement privacy by design in machine learning pipelines, the approach also gives rise to many new research directions and relies on currently open problems.
In this section, we point out a few promising research directions that could help to improve this approach further.

\paragraph{Retrieving relevant documents.}
In theory, access to more documents should improve the results, but selecting the most relevant documents can be challenging in practice.
The integration of access control into the model enables prioritization of the results by considering the access control rights.
For example, a document created by a more trusted or senior user can receive higher priority.

\paragraph{Susceptibility to mislabeling/poisoning.}
So far we have assumed that the security labels are a reliable source of truth.
However, in some applications labelling mistakes may occur either through unintentional mislabels or intentional data poisoning.
After the system has been deployed, an adversary might surreptitiously grant the victim access to poisoned data (\eg, a document containing incorrect information about a specific topic), which the victim's model might use.
This would not be detected by training-time poisoning mitigations (e.g., \cite{wang2019neural}), and so would require new defense techniques.
Nevertheless, this threat can be partially addressed through source attribution that enables users to view the documents used in generating the output and verify their ownership.
Furthermore, it is interesting to note that non-interference can also be used to enforce integrity guarantees.
Data items would have an additional security label conveying information about their integrity (or data provenance) and the user could specify that only data items whose integrity labels satisfy some criteria can be used.
In the same way that non-interference provides confidentiality (see \autoref{sec:non-interference}), it can also be used to ensure integrity based on the integrity labels.

\paragraph{Efficient retrievers.}
Implementing information flow control in machine learning models necessitates preprocessing the input, such as retrieving specific documents and verifying access control rights.
This may result in an increased inference time compared to conventional models.
There has been significant work on developing fast approximate nearest neighbour search.
Developing a system that can take into account access control restrictions would improve inference efficiency and result in a more scalable system.

\paragraph{Access control information as prioritization technique:}
In theory, access to more documents should improve the results, but selecting the most relevant documents can be challenging in practice.
The integration of access control into the model enables prioritization of the results by considering the access control rights.
For example, a document created by a more trusted or senior user can receive higher priority.

\paragraph{User interface}
Consider the scenario that the user intents to share new information, for example from a document which has been authored independently.
The decision to share is made by the author, and involves an implicit declassification.
Making such a decision requires detailed knowledge about where the data has been retrieved from.
In a traditional setting, the author would browse documents manually hence is aware of what sources have been used.
In the above experience, this process may be automated.
Source attribution allows us to point the author to original sources consumed, and hence enable better informed decisions about sharing results.
If certain sources are considered confidential, a properly designed UX may allow the user to explicitly exclude documents, thus avoiding unintentional disclosure.

\section{Conclusion}

In this paper, we have advocate for training privacy-preserving machine learning models on sensitive data by utilizing full contextual metadata such as ownership and access permissions rather than treating machine learning components in isolation.
By borrowing concepts from information flow control theory, we defined stronger privacy guarantees and showed that several existing and popular training pipelines can be expressed using the IFC language.
We have also presented experimental results to not only study the theoretical privacy guarantees but also investigate how existing approaches compare in terms of model utility.
A clear frontrunner has emerged---retrieval augmented architectures.
Retrieval augmented architectures not only have clear advantages in privacy, scalability, and flexibility under label and data updates, but we also observed a significant boost in utility compared to the alternatives.

\bibliographystyle{ACM-Reference-Format}
\bibliography{bibliography}

\appendix

\section{Overview of Retrieval Augmented Architectures}
\label{sec:retrieval_augmentation}
In this section, we introduce different implementations for retrieval augmented architectures in existing literature.
\begin{description}
  \item[Prompting methods]
     Prompting methods are among the most popular approaches to implement retrieval augmentation for text-related tasks. Prompting methods augment the text prompt with retrieved data.
     Prompting-based retrieval augmentation can be easily applied to existing pre-trained models as it typically operates on the model input without altering the pre-existing model architecture. 
     For example, RAG \cite{lewis2020retrieval} and REALM \cite{guu2020realm} applies the prompting method to a transformer-based sequence-to-sequence model or a language model by prepending retrieved data to the original prompt in the form of text tokens.
     Despite its ease of application, prompting methods are restricted by the input length constraints of the underlying pre-trained model.
     Prompting methods can be applied directly to a base model without additional fine-tuning \cite{shi2023replug}, but the extent to which they work depends on the size of the base model. Alternatively, they can also be trained end-to-end for specific tasks.

  \item[Decoding methods]
     Decoding methods conduct retrieval augmentation at the decoding stage. Like prompting techniques, decoding methods operate as a wrapper method for any model without modifying its underlying architecture. For example, kNN-LM \cite{Khandelwal2020Generalization} utilizes retrieved information to interpolate the distribution of the output tokens with a distribution derived from retrieved tokens. 
     KID \cite{liu2022knowledge} constructs a knowledge trie from retrieved documents and employs reinforcement learning to learn a policy to guide the decoder to optimize the knowledge overlap of predicted text with the knowledge trie.
    Although decoding methods do not have a length constraint, they vary in performance, latency, and interpretability depending on how the output distribution was modified.
  
  \item[Cross attention methods]
     Another popular approach for retrieval augmentation is to use the cross attention mechanism to incorporate retrieved information.
     Typical model architectures include RETRO \cite{pmlr-v162-borgeaud22a}, Guided Transformer \cite{hashemi2020guided}, and Memorizing Transformer \cite{wu2022memorizing}. 
     In these architectures, they usually employ a cross attention block to merge the vector representations of original prompt with retrieved data. 
     These approaches also alleviate the length constraint on retrieved information, however integrating them into pre-existing large models is challenging as they require end-to-end training.
     
  \item[Vector fusion methods]
    Vector fusion, which merges vectors from different embedding spaces, is another way to implement retrieval augmentation. This can be achieved through vector concatenation or more sophisticated methods.
    The cross attention methods can also be considered as a type of vector fusion. Although they generally perform better than simple vector fusion, vector fusion methods usually have the advantage of lighter computation and ease of application to multimodal data sources due to late fusion.
    Examples of retrieval augmented models with vector fusion include KIF \cite{fan-etal-2021-augmenting}, which retrieves information from both text and image modalities and merges the data by concatenating the encoded input with a weighted sum of retrieved vector representations.
    Another example is SPALM\cite{yogatama-etal-2021-adaptive}, which employs a gating mechanism to combine encoded input with retrieved vectors. 
    Additionally, FID and its extensions \cite{izacard-grave-2021-leveraging, izacard2021distilling, NEURIPS2021_da3fde15} implement a combination of prompting and vector fusion. 
    They enhance the original prompt with each retrieved document using prompting and then separately encode the augmented prompts with multiple encoders. The encoded vectors are concatenated before they are fed to the decoder.
    
\end{description}

Retrieval augmentation has also gained popularity in the image domain. Depending on the format of the retrieved data (text or image), they can be implemented in the same way as text-based retrieval augmentation \cite{chen2021passage}, or they typically employ vector fusion based methods (including attention), as images are usually represented as vectors.
For example, Re-Imagen \cite{chen2022reimagen} introduces several cross attention blocks to their underlying diffusion model to integrate retrieved image vectors. In another study, kNN-Diffusion \cite{sheynin2023knndiffusion} compares several vector fusion techniques, including self-attention, cross attention and linear projection, to merge retrieved image vectors with the CLIP embedding of the original query and conditions a generative model on the fused embedding to generate images. 

\section{Supplementary Experimental Results and Analysis}
\subsection{Hyperparameter Search for Fine-tuning on the Elsevier dataset}

We fine-tune the GPT-3 model with LoRA on the Elsevier data using a fine-tuning API.
Ideally, we would like to fine-tune GPT-3 on the language modeling task. However, due to the requirement of the fine-tuning API to process data as prompt-completion pairs, we fine-tuned the model for next sentence prediction using the training data.
We started with the recommended values of the hyperparameters and performed hyperparameter search in the order of learning rate multiplier, batch size, and prompt loss weight.
To save computation, we performed most of the hyperparameter search with 10k training data. We split 5\% of the training data as a validation set and monitored training and validation losses for each configuration. 
When tuning the learning rate multiplier, we found that a smaller learning rate works better. Then we tried different batch sizes but increasing the batch size made the losses go up.
We also changed the prompt loss weight, which is a hyperparameter that controls how much the model tries to learn to generate the prompt. 
Increasing the prompt loss weight improved the performance, as it brings the next sentence prediction task closer to a language modeling task.

\begin{table*}[!ht]
  \caption{Hyperparameter search with 10k training data. Note that changing the prompt loss weight also changes the range of the loss.}
  \begin{tabular}{lllll}
  \toprule
      \textbf{learning\_rate\_multiplier} & \textbf{batch\_size} & \textbf{prompt\_loss\_weight} & \textbf{train loss} & \textbf{val loss}  \\ \midrule
      0.02 & 8 & 0.1 & 0.828 & 0.864  \\
      0.05 & 8 & 0.1 & 0.843 & 0.840  \\ 
      0.1 & 8 & 0.1 & 0.864 & went up  \\ 
      0.5 & 8 & 0.1 & went up & went up  \\ 
      2 & 8 & 0.1 & went up & went up  \\ 
      0.05 & 128 & 0.1 & went up & went up  \\ 
      0.05 & 256 & 0.1 & went up & went up  \\ 
      0.05 & 8 & 0.5 & 0.790 & 0.862  \\ 
      0.02 & 8 & 0.5 & 0.803 & 0.812  \\ \bottomrule
  \end{tabular}
\end{table*}

Finally, we fine-tune the model on the full training data with our best hyperparameter setting: learning rate multiplier = 0.05, batch size = 8, prompt loss weight = 0.5, and report the result of the fine-tuned baseline in the paper.

\subsection[Retrieval Augmentation with Different K Values]{Retrieval Augmentation with Different $K$ Values}

Performance of retrieval augmentation depends on size the top $K$ retrieved texts. In our experiments, we experimented with different K values (8, 16, 24) for retrieval. 
As we used the prompting-based approach to implement retrieval augmentation, the maximum $K$ is constrained by the input length limit of the GPT-3 Davinci model (4096 tokens). 
The following tables show the results of retrieval augmentation with different $K$ values. All results are reported on the test set of 1,926 abstracts. 
We find that $K = 24$ yields the best results on the Elsevier dataset and $K = 16$ yields the best result on the Arxiv dataset. 

\begin{table*}[!ht]
  \caption{Results of retrieval augmented models with $K$ = 8}
  \begin{tabular}{lcccccc}
  \toprule
      \textbf{Dataset} & & \textbf{Elsevier} & & & \textbf{Arxiv} & \\
      \textbf{Prompt percentage} & \textbf{10\%} & \textbf{25\%} & \textbf{50\%} & \textbf{10\%} & \textbf{25\%} & \textbf{50\%}  \\ \midrule
      RA: Retrieving from current article & 8.860 & 8.748 & 8.805  & 7.649 & 7.467 & 7.484\\
      RA: Retrieving from random & 11.190 & 10.735 & 10.524  & 13.216 & 12.623 & 12.189 \\
      RA: Access to authors' articles & 9.645 & 9.478 & 9.502 & 8.448 & 8.101 & 7.993\\
      RA: Access to authors' other articles & 10.199 & 9.970 & 9.934  & 10.835 & 10.633 & 10.535\\
      \bottomrule
  \end{tabular}
\end{table*}

\begin{table*}[!ht]
  \caption{Results of retrieval augmented models with $K = 16$}
  \begin{tabular}{lcccccc}
  \toprule
  \textbf{Dataset} & & \textbf{Elsevier} & & & \textbf{Arxiv} & \\
  \textbf{Prompt percentage} & \textbf{10\%} & \textbf{25\%} & \textbf{50\%} & \textbf{10\%} & \textbf{25\%} & \textbf{50\%}  \\ \midrule
      RA: Retrieving from current article & 7.946 & 7.843 & 7.940 & 7.003 & 6.837 & 6.843 \\ 
      RA: Retrieving from random & 11.178 & 10.728 & 10.517 & 13.207 & 12.588 & 12.187 \\
      RA: Access to authors' articles & 9.071 & 8.947 & 9.030 & 7.824 & 7.491 & 7.398  \\ 
      RA: Access to authors' other articles & 9.915 & 9.722 & 9.680 & 10.568 & 10.389 & 10.326 \\ 
      \bottomrule
  \end{tabular}
\end{table*}

\begin{table*}[!ht]
  \caption{Results of retrieval augmented models with $K = 24$}
  \begin{tabular}{lcccccc}
  \toprule
  \textbf{Dataset} & & \textbf{Elsevier} & & & \textbf{Arxiv} & \\
  \textbf{Prompt percentage} & \textbf{10\%} & \textbf{25\%} & \textbf{50\%} & \textbf{10\%} & \textbf{25\%} & \textbf{50\%}  \\ \midrule
      RA: Retrieving from current article & 7.397 & 7.267 & 7.387 & 7.546 & 7.443 & 7.485 \\
      RA: Retrieving from random & 11.184 & 10.728 & 10.514 & 13.205 & 12.601 & 12.195 \\ 
      RA: Access to authors' articles & 8.599 & 7.781 & 7.811 & 8.351 & 8.101 & 8.074 \\
      RA: Access to authors' other articles & 9.706 & 9.503 & 9.500 & 10.729 & 10.564 & 10.488 \\
      \bottomrule
  \end{tabular}
\end{table*}

\subsection{Model Utility}
\label{sec:utility}
\begin{figure*}[!ht]
  \centering
  \begin{tikzpicture}
 
\begin{axis} [ybar = .05cm,
    bar width = 2pt,
    legend pos=outer north east,
    legend style={font=\small},
    xtick={0,1,...,10},
    ylabel = Perplexity
]
 
\addplot table [x=percentile, y=gpt3, col sep=comma] {result/plot_data.csv};
\addplot table [x=percentile, y=access to current, col sep=comma] {result/plot_data.csv};
\addplot table [x=percentile, y=access to collection, col sep=comma] {result/plot_data.csv};
\addplot table [x=percentile, y=access to other, col sep=comma] {result/plot_data.csv};
\addplot table [x=percentile, y=access to random, col sep=comma] {result/plot_data.csv};

\legend{GPT-3 zero-shot, Access to current, Access to collection, Access to other articles, Access to random}
\end{axis}
 
\end{tikzpicture}
  \caption{Perplexity of the test set divided into 10 percentile groups by the zero-shot baseline, and improvements in perplexity achieved by retrieval augmented models on each percentile group.}
  \label{fig:perplexity}
  \Description{}
\end{figure*}
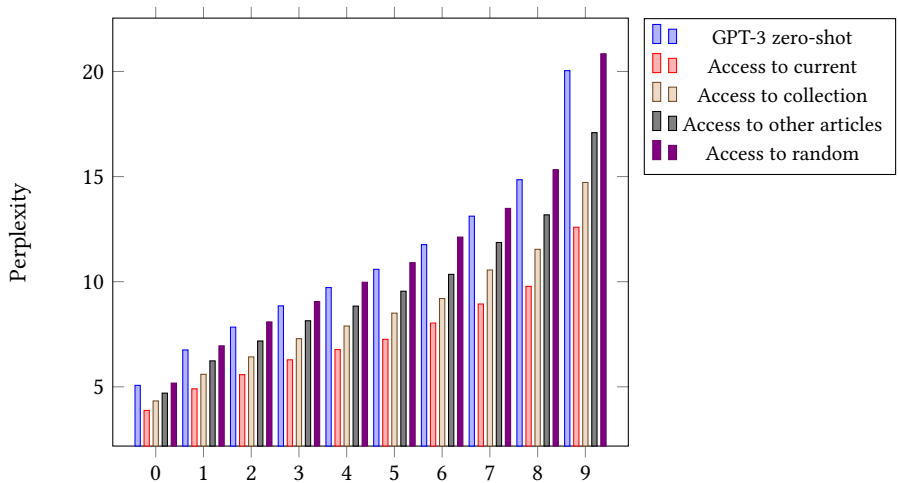

In this section, we perform further analysis on the utility of the retrieval augmented models. As the average perplexity of the abstracts by the GPT-3 zero-shot baseline is already quite low, to better analyse the utility of our proposed retrieval-augmented models, we divide the test set into 10 percentile groups based on the perplexity scores from the GPT-3 zero-shot baseline. 
\autoref{fig:perplexity} compares the performance of the retrieval augmented models against the zero-shot baseline across each of the percentiles from a single run of our experiments.
The figure demonstrates that, consistent with our previous observations, having access to more relevant data groups in retrieval augmentation results in more pronounced improvements in text prediction across all percentiles. 
In addition, the figure also reveals a general trend across the percentiles that retrieval augmentation has a greater impact on cases where the baseline perplexity is higher (as one moves towards the right side of the figure). 
This suggests that retrieval augmentation is more beneficial for predicting text that the baseline model was uncertain about.

\begin{figure*}[!ht]
	\centering
	\includegraphics[width=\textwidth, trim=0 25cm 0 0]{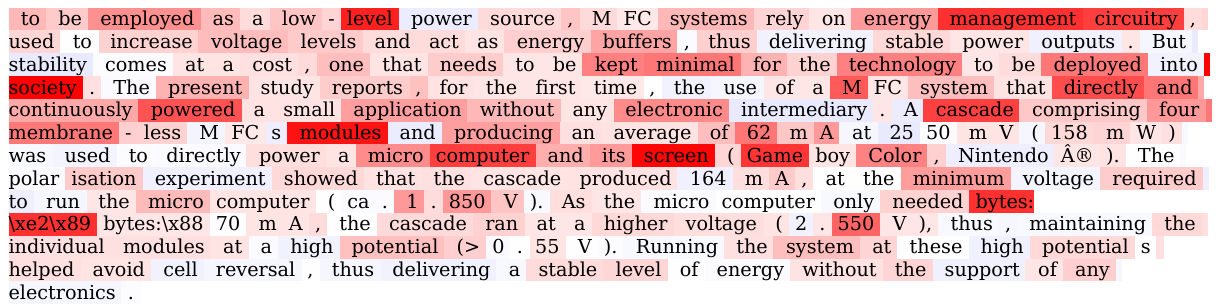}
	\caption{The effect of retrieval augmentation on the token probabilities of a random abstract. The intensity of color denotes the relative changes of the token probabilities, with red showing an increase, and blue showing a decrease.}
	\label{fig:colormap}
  \Description{}
\end{figure*}

\begin{table}[!ht]
  \caption{Top 4 (out of K = 24) retrieved texts for the given prompt.}
  \centering
	\begin{tabular}{lp{0.8\linewidth}}
		\toprule
		\textbf{Prompt} & \textit{Many studies have demonstrated that microbial fuel cells (MFC) can be energy-positive systems and power various low power applications. However,} \\ \midrule
		\textbf{Retrieved} &  Research on the topic has focussed on microbial fuel cells (MFC) both as a remediation technology for \textbf{wastewater treatment} and as a \textbf{low-level renewable energy source} for applications \\
		& The results presented here demonstrate for the first time that a \textbf{microcomputer} can \textbf{directly and continuously} be powered by MFCs \textbf{without any energy management apparatus}. \\
		& Although the results demonstrated that a MFC system can \textbf{directly and continuously power a microcomputer}, there is room for improvement, especially on scaling the system to the end application. \\
		& The objective was to demonstrate that MFCs could directly and continuously power an exemplar \textbf{microcomputer and its screen} without the use of any \textbf{power management electronic circuitry}. \\
		\bottomrule
	\end{tabular}
	\label{tab:ra_example}
\end{table}

\autoref{fig:colormap} presents a visualization of tokens in a randomly chosen abstract and the effect of retrieval augmentation on the token probabilities.
\autoref{tab:ra_example} shows the top 4 (out of $K = 24$) retrieved texts for the given prompt of the abstract, and we highlight phrases that are relevant to the generation of the abstract. 
The complete list of the top $K = 24$ retrieved texts for this example can be found in the end of this section. 
In this example, it is evident that retrieval augmentation generally enhances the probability estimates for the majority of the tokens in the abstract. 
Furthermore, we can observe that the effect of retrieval augmentation is not uniform throughout the abstract. 
By comparing the phrases with the greatest improvement against the retrieved text, it becomes clear that the retrieval augmented model effectively leveraged the retrieved data to predict phrases that may be infrequent in everyday language but present in the retrieved data.

For reference, the complete list of the top 24 retrieved texts for the given example is given below:

\textbf{Prompt}

\textit{Many studies have demonstrated that microbial fuel cells (MFC) can be energy-positive systems and power various low power applications. However,}

\textbf{Top 24 retrieved texts}

Text: Research on the topic has focussed on microbial fuel cells (MFC) both as a remediation technology for wastewater treatment and as a low-level renewable energy source for applications.

Text: The results presented here demonstrate for the first time that a microcomputer can directly and continuously be powered by MFCs without any energy management apparatus.

Text: Although the results demonstrated that a MFC system can directly and continuously power a microcomputer, there is room for improvement, especially on scaling the system to the end application.

Text: The objective was to demonstrate that MFCs could directly and continuously power an exemplar microcomputer and its screen without the use of any power management electronic circuitry.

Text: This showed that the system had to be calibrated to meet the application-specific needs, which in this case was a voltage higher than 1.832 V. Due to its modular nature, the MFC system can be adapted to any scale of use and yet be energy-positive, which is a feature not always found in other biotechnologies (e.g . biogas).

Text: Hence, all reports that have demonstrably powered applications with MFCs used power management circuitry.

Text: Therefore, the objective here was to investigate if MFCs could directly power an off-the-shelf application without the need for any electronic circuitry.

Text: Employing power management systems has been a solution exploited so far to remediate the intrinsic current instability and low-power density of microbial fuel cells.

Text: To be used as an energy source Microbial fuel cells are often seen as too unstable and incapable of producing directly exploitable energy levels.

Text: These two factors, which are inherent to the technology, have driven the development of MFC-systems aimed at powering real-world applications that need high power densities.

Text: This is because the MFC technology is the only BES that can directly convert wastewater into electrical energy [3–5], thus saving energy for a process that normally consumes a lot of electrical power.

Text: Several studies have shown that MFC systems can power various applications, without voltage reversal, through the use of either DC/DC converter [14–19] or capacitor banks combined with purpose-built management circuitry [4,20,21].

Text: However, as pointed out by Chen et al . [12], the cost of MFC systems has to be kept minimal for technology to be deployed in society, and power management systems represent an additional cost that could limit the technology's implementation.

Text: A key aspect that would require attention, especially in terms of employing MFCs as a power source, is to explore the robustness of such system over a long period of time and investigate the stability of the series electrical connection.

Text: Since the optimal operating potential for the S-MFC modules is around 450 mV [17,18], four modules electrically connected in series should have the potential to directly power the GBC.

Text: While research on MFC was first reported in 1911 [1], bioelectrochemical systems (BES) converting the chemical energy contained in reduced organic matter into electrical energy have recently received great attention, as illustrated by the number of publications on the subject [2].

Text: In addition, due to diffusion limitations and increased internal resistance, smaller MFCs are more power dense than larger ones [8].

Text: Hence, the hypothesis was to employ relatively large size MFC modules, with respect to the target application, and operating the system under suboptimal feeding conditions, to simulate real or field environments and test the system capability in terms of voltage reversal.

Text: At single MFC level, the maximum power transfer point is usually occurring at a voltage between 300 mV and 500 mV, depending on the design and the material employed.

Text: The 4-module S-MFC cascade continuously produced 2.55 V and 61 mA, a total of 130 mW, which is above the minimum requirements of the microcomputer and its screen (GBC).

Text: This type of design exploits the capacity of complex microbial consortia to structure themselves and the environment in which they thrive in a succession of horizontal layers following a redox gradient: the more reduced layers are at the bottom of the electrolyte column and the more oxidised layers are at the top of the column.

Text: Following our past experience in powering application [17,18], the stack assembled here employed self-stratifying membraneless MFCs modules (S-MFCs).

Text: The four S-MFC modules employed in the present report were designed following the results from previous studies investigating scalability, materials and functional parameters of urine-fed S-MFCs [22–24].

Text: Based on the results above, the hypothesis was that this cascade could power the GBC directly without the use of any power management system or energy buffers such as batteries and/or capacitors (i.e . directly).

\end{document}